\documentclass[11pt,a4paper]{article}
\usepackage[hyperref]{acl2021}
\usepackage{times}
\usepackage{latexsym}

\usepackage{array}
\usepackage{graphicx}
\usepackage{tabularx,tabulary}
\usepackage{url}
\usepackage{nicefrac}
\usepackage{calc}
\usepackage{xcolor}
\usepackage{graphicx}
\usepackage{multirow}

\usepackage{amsthm}
\usepackage{amssymb}
\usepackage{amsmath}
\usepackage{amsfonts}
\usepackage[toc]{appendix}
\usepackage{color, colortbl}
\usepackage{adjustbox}
\usepackage{tabularx}
\usepackage{booktabs}
\usepackage{subcaption}
\usepackage{makecell}
\usepackage{arydshln}
\usepackage{tikz}
\usepackage{enumitem}

\usepackage{microtype}

\aclfinalcopy 

\title{\textsc{GenSF}: Simultaneous Adaptation of Generative Pre-trained Models and Slot Filling } 
\author{Shikib Mehri \and Maxine Eskenazi \\
  Language Technologies Institute, Carnegie Mellon University \\
  \texttt{\{amehri,max\}@cs.cmu.edu}}

\date{}

\begin{document}
\maketitle
\begin{abstract}

In transfer learning, it is imperative to achieve strong alignment between a pre-trained model and a downstream task. Prior work has done this by proposing task-specific pre-training objectives, which sacrifices the inherent scalability of the transfer learning paradigm. We instead achieve strong alignment by simultaneously modifying both the pre-trained model and the formulation of the downstream task, which is more efficient and preserves the scalability of transfer learning. We present \textsc{GenSF} (\textbf{Gen}erative \textbf{S}lot \textbf{F}illing), which leverages a generative pre-trained open-domain dialog model for slot filling. \textsc{GenSF} (1) adapts the pre-trained model by incorporating inductive biases about the task and (2) adapts the downstream task by reformulating slot filling to better leverage the pre-trained model's capabilities. \textsc{GenSF} achieves state-of-the-art results on two slot filling datasets with strong gains in few-shot and zero-shot settings. We achieve a \textbf{9 $\mathbf{F_1}$ score} improvement in zero-shot slot filling. This highlights the value of strong alignment between the pre-trained model and the downstream task.

\end{abstract}

\section{Introduction}

The advent of pre-trained language models \citep{devlin-etal-2019-bert,radford2019language} has transformed natural language processing. The dominant paradigm has shifted away from designing task-specific architectures towards transfer learning. Fine-tuning pre-trained models on downstream datasets achieves strong performance on a variety of natural language understanding tasks \citep{wang2018glue}. Generally, prior to fine-tuning, the pre-trained models are adapted to the specifics of the downstream task through minor architectural modifications (e.g., adding a classification layer) \citep{chen2019bert,mehri2020dialoglue}. By avoiding major task-specific changes to the models, it is assumed that the underlying pre-trained models possess a degree of generality that allows transfer to a variety of tasks. We posit that this assumption is flawed. Consequently this paper demonstrates the importance of incorporating \textit{inductive biases} that achieve \textit{stronger alignment} between the pre-trained model and the downstream task.

\begin{figure}
    \centering
    \includegraphics[width=\linewidth]{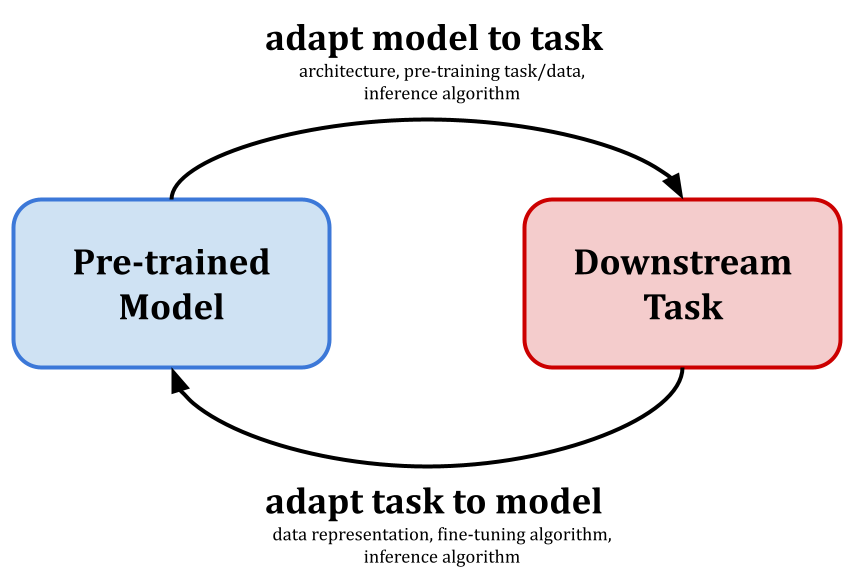}
    \caption{To achieve a stronger alignment, both the downstream task and the pre-trained models must be adapted. The downstream task can be adapted with knowledge of the properties and capabilities of the pre-trained models. Likewise, the pre-trained model can be adapted with knowledge of the downstream task/data.}
    \label{fig:my_label}
\end{figure}

Recent work has validated the idea that stronger alignment between pre-training and the downstream task results in improved performance. Rather than fine-tuning off-the-shelf models, it is more effective to first understand the downstream task and adapt the model's architecture, pre-training and inference algorithm accordingly. Adapting pre-trained models in this manner is equivalent to \textbf{incorporating inductive biases about the downstream task}. For example, pre-training on open-domain dialog data results improves performance on downstream dialog tasks \citep{henderson2019convert,mehri2020dialoglue}. Designing task-specific pre-training objectives has yielded strong results in extractive question answering \citep{glass2019span}, paraphrase and translation \citep{lewis2020pre} and slot filling \citep{henderson2020convex}. This body of work attains stronger alignment by significantly modifying the pre-trained model through task-specific pre-training. This necessitates a new pre-trained model for every downstream task, and therefore relinquishes the inherent scalability of the transfer learning paradigm. Instead, we  achieve stronger alignment by \textit{simultaneously adapting} both the pre-trained model and the downstream task, such that both contain inductive biases about one another.

The downstream task can be adapted to achieve stronger alignment with the capabilities of the pre-trained model. To effectively leverage pre-trained models, it is important to first understand the properties and capabilities of the model derived from the model architecture, the pre-training data and task. Then the downstream task can be adapted to be better aligned with the model. Adapting the task to the model is equivalent to \textbf{incorporating inductive biases about the pre-trained model} into the downstream task. For example, given a pre-trained model that was trained with a ranking objective, it is likely to be more effective if the downstream fine-tuning and inference algorithms are modified to rank rather than to classify. By simultaneously adapting both the downstream task and the pre-trained model, we intend to achieve stronger alignment without sacrificing the inherent scalability of the transfer learning paradigm (i.e., avoiding task-specific pre-trained models).

We address the task of slot filling, a natural language understanding task with the goal of identifying values for pre-defined attributes (slots) in a natural language utterance. We leverage a DialoGPT \citep{zhang-etal-2020-dialogpt}, a generative language model, pre-trained on open-domain dialog data. To achieve strong alignment between the slot filling task and DialoGPT, we (1) reformulate slot filling as a natural language response generation task, and (2) augment the DialoGPT architecture with a copy-mechanism, constrained decoding and a post-processing heuristic. The resulting model, \textsc{GenSF} (\textbf{Gen}erative \textbf{S}lot \textbf{F}illing), is shown to achieve state-of-the-art results on two slot filling datasets. \textsc{GenSF} achieves the strongest performance gains in few-shot and zero-shot settings, highlighting the importance of stronger alignment in the absence of abundant data. Our code is open-sourced and can be found at \textbf{\url{https://github.com/shikib/generative_slot_filling}}.

\section{Related Work}

Slot filling is the task of identifying values for pre-defined attributes, or slots, in a natural language utterance \citep{tur2011spoken}. Slot filling is a vital natural language understanding component of task-oriented dialog systems \citep{young2002talking,young2010still}. A variety of architectures have been explored for the task of slot filling, including CNNs \citep{vu2016sequential}, deep LSTMs \citep{yao2014spoken}, RNNs with external memory \citep{peng2015recurrent}, encoder labeler LSTMs \citep{kurata2016leveraging} and joint pointer and attention seq2seq networks \citep{zhao2018improving}. With the introduction of large-scale pre-trained language models \citep{devlin-etal-2019-bert,radford2019language}, strong slot filling results have been achieved with simple architectures \citep{chen2019bert}. 

Several approaches have been proposed for zero-shot slot filling. \citet{bapna2017towards} leverage slot names and descriptions to align slots across domains. \citet{shah2019robust} leverage examples for zero-shot slot filling. \citet{liu2020coach} achieve strong results in zero-shot slot filling with a coarse-to-fine approach in combination with template regularization. We use the Coach+TR model \citep{liu2020coach} as a baseline in our zero-shot experiments.

Working on the hypothesis that pre-trained language models, such as BERT \citep{devlin-etal-2019-bert}, do not effectively capture the intricacies of dialog, recent work has attempted to mitigate this issue. \citet{coope-etal-2020-span} use ConveRT \citep{henderson2019convert}, a lightweight model pre-trained on dialog data, in combination with CNN and conditional random field (CRF) to outperform BERT. \citet{mehri2020dialoglue} achieves similar results with ConvBERT, a model that further pre-trains BERT on open-domain dialog data. Recently, \citet{henderson2020convex} introduces a \textit{`pairwise cloze'} pre-training objective that uses open-domain dialog data to specifically pre-train for the task of slot filling. The resulting ConVEx model achieves significant improvements, particularly in few-shot settings. A common theme in recent work is achieving better alignment between the pre-trained models and the downstream task, either by pre-training on data that is closer to the domain of the downstream task (i.e., dialog data) \citep{henderson2019convert,mehri2020dialoglue} or by designing custom pre-training objectives that better model the downstream task \citep{henderson2020convex}. Our proposed approach shares the goal of achieving better alignment, but we simultaneously adapt both the pre-trained model and the downstream task, with the goal of leveraging a generative pre-trained dialog model, DialoGPT, for slot filling.

\section{Methods}

In order to effectively leverage a pre-trained generative dialog model, DialoGPT \citep{zhang-etal-2020-dialogpt}, for the task of slot-filling, we introduce the \textsc{GenSF} model which achieves stronger alignment between the downstream task and the pre-trained model, by simultaneously adapting the task to the model and the model to the task. This paper first describes how the slot filling task is reformulated as a natural language response generation task to be better aligned with the DialoGPT model. Next, it describes several modifications to the DialoGPT architecture and inference algorithm that act as inductive biases for the slot filling task. 

\subsection{Slot Filling as Response Generation}

Given an utterance $u = \{ w_1, w_2, \dots w_n \}$, a set of possible slot keys $s = \{s_1, s_2, \dots s_k\}$, and a list of slots requested by the system $r = \{ r_1, r_2, \dots r_m \}$ (where $r_i \in s$ and $m \geq 0$), the task of slot filling is to assign a value to a subset of the slot keys. Concretely, for a given slot key $s_i$, the output will either be \texttt{NULL} or a contiguous span of words from the utterance: $s_i = \{ w_i, \dots w_{i+j} \}$.

In response generation, given a dialog context consisting of a sequence of utterances: $c = \{ x_1, x_2, \dots x_n \}$ wherein each utterance $x_i$ is a sequence of words, the task is to generate a valid response $y = \{w_1, w_2, \dots w_m \}$. 

Many tasks can be represented as an \textit{input to output} mapping \citep{raffel2019exploring,hosseini2020simple,peng2020soloist}, making sequence-to-sequence a universal formulation. Trivially, slot filling can be represented as a sequence-to-sequence task by setting the context to be the concatenation of the utterance and the requested slots: $c = \{ u, r \}$ and the target response to be the slot mappings $y = \{ (s_1, w_{i:j}), (s_2, {\tt NULL}), \dots (s_k, (w_{j:n}) \}$. However, this does not leverage the natural language capabilities of pre-trained dialog models. While this trivial formulation may suffice with sufficient training, it will under-perform in few-shot and zero-shot settings. To this end, this paper presents a reformulation of slot filling that better aligns with the natural language capabilities of DialoGPT.

We hypothesize that to some degree, large-scale dialog pre-training can result in a model implicitly learning to fill slots. For example, given the slot key \textit{`time'}, such a model should understand what \textit{time} is and should be able to generate a valid time (e.g., \textit{`4:15 pm'}). An effective task formulation can leverage these implicitly learned slot filling capabilities. An off-the-shelf pre-trained model is likely to only be capable of filling generic slots (e.g., time, date, price, etc.). But by reformulating slot filling in a manner that is better aligned with the pre-training task, it should be easier for the model to adapt to novel slot keys. 

Concretely, given a slot filling input ($u$, $r$) and a particular slot key $s_i$, we construct a natural language dialog context using a template-based approach: $c =$ \textit{`What is the $ \{ f(r) \} $? \texttt{[eos]} $\{u\}$ \texttt{[eos]} Ok, the $\{f(s_i)\}$ is'}. Here, $f$ denotes a manually constructed function that maps slot keys to a natural language phrase (e.g., $first\_name$: \textit{first name}, $departure\_location$: \textit{leaving from}). Given the constructed dialog context, the model is tasked with completing the partial response (i.e., \textit{Ok, the $\{f(s_i)\}$ is}) by auto-regressively generating the slot value. During training the model would be tasked with generating either the slot value or the phrase \textit{not provided}. With this natural language reformulation, the slot filling task is being adapted to better leverage the capabilities of the pre-trained DialoGPT model. As this achieves better alignment between the pre-trained model and the downstream task, it should be more effective for zero-shot and few-slot filling. To better illustrate the conversion of the slot-filling input (utterance $u$ and request slots $r$), several examples are shown in Table \ref{tab:examples}.

\begin{table*}[]
\newcolumntype{x}[1]{>{\centering\arraybackslash\hspace{0pt}}p{#1}}
\renewcommand*{\arraystretch}{1.3}

    \centering
    \begin{tabular}{|>{\centering\arraybackslash} m{0.25\linewidth}|>{\centering\arraybackslash} m{0.18\linewidth}|>{\centering\arraybackslash} m{0.15\linewidth}|>{\centering\arraybackslash} m{0.3\linewidth}|}
    \hline
        \textbf{Utterance} & \textbf{Requested Slots} & \textbf{Slot Key} & \textbf{Natural Language Context}   \\ \hline
        We will require an outside table to seat 9 people on August 23rd & None & \texttt{date} &  We will require an outside table to seat 9 people on August 23rd \texttt{[EOS]} Ok, the date is \\ \hline
        
        Laurice Hoisl & \texttt{first\_name}, \texttt{last\_name} & \texttt{first\_name} & What is the first name, last name? \texttt{[EOS]} Laurice Hoisl \texttt{[EOS]} Ok, the first name is \\ \hline  
        My party will be 9 people. My name is Nancie Waltemeyer and the time is 7pm & None & \texttt{people} & My party will be 9 people. My name is Nancie Waltemeyer and the time is 7pm \texttt{[EOS]} Ok, the number of people is \\ \hline
    \end{tabular}
    \caption{Examples of slot filling inputs reformulated as natural language dialog contexts}
    \label{tab:examples}
\end{table*}

\subsection{DialoGPT for Slot Filling}

In order to adapt the pre-trained DialoGPT model to the slot filling task, we augment the architecture and modify the inference algorithm. These adaptations are motivated by the observation that if the slot value is provided, it will always be a contiguous span of tokens from the utterance. As such, the generative model can only produce: (1) \textit{`not provided`} if the slot does not appear in the utterance, (2) the end of sentence token, and (3) tokens from the input utterance.

A copy-mechanism is incorporated into the DialoGPT architecture to allow the model to explicitly generate tokens from the input utterance. Given a context $c = \{x_1, x_2, \dots x_n\}$, through its self-attention layers, the model will produce a hidden state representation for each token, $h = \{h_1, h_2, \dots, h_n \}$. A probability distribution over the vocabulary is then obtained by passing $h_n$ through a classification layer:

\begin{equation}
P_{vocab} = \textrm{softmax}(Wh_n + b)
\end{equation}

To explicitly generate tokens from the input, $h_n$ is used to attend to $h_{1:n}$ to produce a probability distribution over $x_{1:n}$. The process for computing the probability for a specific word, $P_{copy}(w)$ is as follows:

\begin{equation}
    \alpha = \textrm{softmax}(h_n^T  h_{1:n})
\end{equation}

\begin{equation}
    P_{copy}(w) = \sum_{i:x_i = w} \alpha_i
\end{equation}

These two probability distributions are combined through a weighted sum. The weight assigned to each of the distributions is predicted using $h_n$:

\begin{equation}
    p_{copy} = \sigma ( W_{copy} h_n + b_{copy} )
\end{equation}

The final probability distribution is therefore:

\begin{equation}
    P_{final} = (1 - p_{copy}) P_{vocab} + p_{copy} P_{copy}
\end{equation}

The copy-mechanism requires training, as it introduces new weights ($w_{copy}$, $b_{copy}$) and the off-the-shelf DialoGPT model does not necessarily produce attention weights, $\alpha$, that can be used to create an output probability distribution. As such, to attain strong zero-shot performance we must also modify the inference algorithm to account for the aforementioned observation. This is done using both constrained decoding and a post-processing heuristic.

Constrained decoding is a modification of greedy decoding wherein the argmax sampling is modified to only generate (1) words that appear in the input utterance, (2) the end of sentence token and (3) the phrase \textit{`not provided'}. 

The slot values may consist of terms that the model has not frequently observed during pre-training (e.g., names, times). As such, because the DialoGPT model leverages a subword vocabulary, some subword tokens may be dropped during generation and therefore the slot values may be generated with typos (e.g., `Mocer' vs `Mocher'). A simple post-processing heuristic is applied to mitigate this problem. If the slot value produced by the model is not present in the utterance, the Levenshtein distance to every contiguous span of tokens in the utterance is computed. If the best edit distance is within a certain threshold ($0.3 \times \textrm{len}(y)$), the corresponding span is returned as the slot value.

Through these modifications, the DialoGPT model is adapted to reflect the properties of the slot filling task. The copy-mechanism, constrained decoding and post-processing mechanism serve as an inductive bias to enable the pre-trained model to be better adapted for the downstream slot filling task.
\begin{table*}[]
\small
\renewcommand*{\arraystretch}{1.1}

    \centering
{%
\begin{tabularx}{0.7\textwidth}{@{\extracolsep{\stretch{1}}} c *{4}{c}}
\toprule
\textbf{Fraction} & \textbf{Span-ConveRT}  & \textbf{Span-BERT} &  \textbf{ConVEx} & \textbf{GenSF} \\ \midrule
1 (8198)        & {95.8}              & 93.1  &  96.0 & \textbf{96.1} \\
$\nicefrac{1}{2}$ (4099)     & 94.1                & 91.4 &  94.1 & \textbf{94.3} \\
$\nicefrac{1}{4}$ (2049)     & {91.2}              & 88.0 &  92.6 & \textbf{93.2}      \\
$\nicefrac{1}{8}$ (1024)     &{88.5}             & 85.3  &  90.6  & \textbf{91.8}   \\
$\nicefrac{1}{16}$ (512)     & {81.1}            & 76.6 &   86.4 & \textbf{89.7}      \\
$\nicefrac{1}{32}$ (256)     &{63.8}               & 53.6  & 81.8 & \textbf{82.1}     \\
$\nicefrac{1}{64}$ (128)     & {57.6}              & 42.2&  \textbf{76.0} & \textbf{76.1}      \\
$\nicefrac{1}{128}$ (64)    & {40.5}              & 30.6 &   71.7 & \textbf{72.2}      \\
\bottomrule
\end{tabularx}
}
    \caption{$F_1$ scores across all slots for the evaluation on the \textsc{restaurants-8k} test data with varying proportions of the training set. Numbers in brackets denote the training set sizes. The best scores (statistically significant by t-test to $p < 0.05$) are shown in boldface.}
    \label{tab:r8k_results}
\end{table*}
\section{Experiments}

Experiments are performed to empirically validate the hypothesis that simultaneously adapting the downstream task and the pre-trained model results in stronger alignment and improved performance. We present experiments on two datasets and assess \textsc{GenSF} in full-data, few-shot and zero-shot settings. An ablation study is performed to characterize the source of the performance gains and demonstrate the importance of simultaneous adaptation.

\subsection{Datasets}

Experiments are carried out on \textsc{restaurants-8k} \citep{coope-etal-2020-span} and the \textsc{dstc8} datasets \citep{rastogi2020schema}. \textsc{restaurants-8k} consists of 8,198 utterances from a commercial restaurant booking system and includes 5 slots (date, time, people, first name, last name). The \textsc{dstc8} datasets span four different domains (buses, events, homes, rental cars) for a total of 5,569 utterances with slot annotations extracted by \citet{coope-etal-2020-span}.

In both datasets, the value for a particular slot is always a contiguous span of the utterance. Some utterances consist of a set of slots requested by the system prior to the user utterance. This allows an otherwise ambiguous utterance like \textit{`four'} to be interpreted as either \textit{`four people'} or \textit{`four o'clock'}.

\subsection{Experimental Setup}

We use the pre-processing and evaluation scripts provided by the DialoGLUE benchmark \citep{mehri2020dialoglue}. We follow the setup of \citet{coope-etal-2020-span} and \citet{henderson2020convex}, wherein a validation set is not used and the experiments are therefore performed with fixed hyperparameters. Throughout all the experiments, the medium version of DialoGPT \citep{zhang-etal-2020-dialogpt} is used. We use the AdamW optimizer \citep{loshchilov2017decoupled} with a learning rate of \texttt{5e-5}. On \textsc{restaurants-8k}, the models are trained for 10 epochs in the full-data setting, 20 epochs in the few-shot settings and 40 epochs in the extreme few-shot settings ($\nicefrac{1}{32}$ - $\nicefrac{1}{128}$; or less than $256$ training examples). On the \textsc{dstc8} datasets, the models are trained for 20 epochs in the full-data setting and 40 epochs in the few-shot setting.

The models are evaluated on the full test set, regardless of the amount of training data, using macro-averaged $F_1$ score \citep{coope-etal-2020-span}.

To facilitate reproducibility, the code and the trained models will be released upon publication.

\begin{table*}[]
\small
\renewcommand*{\arraystretch}{1.1}

    \centering
{%
\begin{tabularx}{0.8\textwidth}{@{\extracolsep{\stretch{1}}} c *{5}{c}}
\toprule

       & \textbf{Setting} & \textbf{Span-ConveRT} & \textbf{Span-BERT} &  \textbf{ConVEx} & \textbf{GenSF}\\ \midrule
Buses\_1  
& Full-Data (1133)       & {93.5}                 & 93.3    & 96.0 & \textbf{98.1} \\
              & Few-Shot (283)     & {84.0}                 & 77.8  & {86.7} & \textbf{90.5}    \\ \midrule
Events\_1 
& Full-Data (1498)       & 92.7                 & 84.3  & 91.7 & \textbf{94.7}  \\
              & Few-Shot (374)      & {82.2}                & 78.6  &  87.2 & \textbf{91.2}    \\ \midrule
Homes\_1   
              & Full-Data (2064)                    &{ 94.8 }             & {96.3}  & \textbf{98.3}  & 96.9   \\
              & Few-Shot (516)     & \textbf{95.4}                & \textbf{95.1} &   {94.5}    & 93.7  \\ \midrule
RentalCars\_1 
& Full-Data (874)                                  & \textbf{94.0}              & 92.8 & {92.0} & 93.5     \\
              & Few-Shot (218)     & {83.0}                & {81.4} & \textbf{87.4} & 86.7    \\
              \bottomrule

\end{tabularx}
}
    \caption{$F_1$ scores across all slots for evaluation on the \textsc{dstc8} single-domain datasets in the full-data and few-shot settings. Numbers in brackets denote training set sizes. The best scores (statistically significant by t-test, to $p < 0.05$) are shown in boldface.}
    \label{tab:dstc8_results}
\end{table*}

\subsection{Slot Filling Results}

Throughout the experiments we compare to several models from prior work. Span-ConveRT \citep{coope-etal-2020-span} and Span-BERT train a CNN and a CRF on top of contextual subword embeddings produced by ConveRT \citep{henderson-etal-2020-convert} and BERT \citep{devlin-etal-2019-bert}, respectively. ConVEx \citep{henderson2020convex} devises a \textit{pairwise cloze} pre-training objective specifically for slot-filling. This task-specific pre-training objective is an example of significantly adapting the pre-trained model to the downstream task. In contrast to ConVEx, \textsc{GenSF} achieves strong alignment between the pre-trained model and the downstream task by simultaneously adapting both the task and the model. As such, \textsc{GenSF} does not need a task-specific pre-trained model and is inherently more scalable. The ConVEx pre-training takes 8 hours to train on 12 GPUs, while \textsc{GenSF} takes less than four hours to train on a single GTX 1080TI.

As shown in Table \ref{tab:r8k_results}, \textsc{GenSF} achieves state-of-the-art results across all experimental settings on the \textsc{restaurants-8k} dataset. In the full-data setting, \textsc{GenSF} slightly outperforms ConVEx. Though the performance gain is small, this result signifies that our model can leverage an abundance of data. The value of strong alignment between the downstream task and the pre-trained model is better exemplified in the few-shot settings. Especially in the extreme few-shot settings (i.e., $\nicefrac{1}{32}$ - $\nicefrac{1}{128}$ of the training set), \textsc{GenSF} strongly outperforms Span-ConveRT and Span-BERT, with greater than $30$ $F_1$ score improvements. The few-shot performance of both ConVEx and \textsc{GenSF} in these few-shot settings underlies the value of effectively aligning the pre-trained model and the downstream task. However, \textsc{GenSF} achieves this alignment by simultaneously incorporating inductive biases about the model into the task rather than designing a complex pre-training objective. By incorporating inductive biases into both the task and the model, the approach outlined in this paper does not require task-specific pre-trained models and therefore preserves the inherent generality of the transfer learning paradigm. Furthermore, \textsc{GenSF} attains moderate improvements over ConVEx, especially in the few-shot settings, with a $3$ $F_1$ score improvement in the $\nicefrac{1}{16}$th setting.

The results on the \textsc{dstc8} single-domain datasets is shown in Table \ref{tab:dstc8_results}. Here, we evaluate on both full-data and few-shot (25\% of the training data) settings. On average, \textsc{GenSF} achieves strong performance improvements over prior work. In the full-data settings the best performance is observed on the buses and events domains, where \textsc{GenSF} achieves a $2.1$ and $3.0$ $F_1$ score improvement over ConVEx, respectively. In the few-shot settings, \textsc{GenSF} achieves a $4.0$ $F_1$ score improvement over ConVEx on these domains and a $6.5$ and $9.0$ point improvement over Span-ConveRT. These strong improvements, over both Span-ConveRT and ConVEx, highlight the value of strong alignment between the pre-trained model and the downstream task, particularly in the few-shot experiments.

\textsc{GenSF} moderately underperforms on the homes and rental cars domains. On the homes domain, \textsc{GenSF} outperforms Span-ConveRT and Span-BERT but scores $1.4$ points below ConVEx. Similarly, on the rental cars domain, \textsc{GenSF} outperforms ConVEx and Span-BERT, but is $0.5$ points below Span-ConveRT. Though \textsc{GenSF} is still competitive in these domains, these results nonetheless highlight a weakness of our model. Our use of a generative pre-trained dialog model, specifically DialoGPT \citep{zhang-etal-2020-dialogpt}, was motivated by the hypothesis that such models can implicitly learn to identify certain slots through response generation pre-training. This hypothesis is empirically validated through improved performance on \textsc{restaurants-8k} and the buses/events domains of \textsc{dstc8}. \textsc{GenSF} relies on the pre-trained model having an implicit understanding of the slots. This implicit understanding results in strong performance on slots like \textit{`time'} or \textit{`first name'}, since such terms are likely to have been observed during pre-training. However, this is not the case for all slots and \textsc{GenSF} can underperform on slots that are ambiguous, ill-defined or are unlikely to have been observed during open-domain dialog pre-training. The homes domain consists of the slot, \textit{`area'}, which has several definitions and is therefore challenging for the pre-trained model to understand and detect. The rental cars domain contains the slots \textit{`pickup date'} and \textit{`dropoff date'}. While the DialoGPT model has learned to detect a \textit{`date'}, the distinction between these two slots is more nuanced and therefore may cause some amount of confusion. As such, while \textsc{GenSF} is competitive in these domains and is only outperformed by one of the three models, these domains demonstrate that there are limitations at present to leveraging a generative pre-trained model. However, it is possible that by further adapting the downstream task to the pre-trained model, for example by renaming these slots (e.g., \textit{`area'} may be renamed to \textit{`city'}), the performance drops may be mitigated. 

Overall, \textsc{GenSF} achieves impressive performance gains in both full-data and few-shot settings, underlying the value of achieving strong alignment between the pre-trained model and the downstream task. Furthermore, \textsc{GenSF} achieves this alignment by simultaneously adapting both the task and the model and without sacrificing the inherent scalability of the transfer learning paradigm or necessitating task-specific pre-training. In the \textsc{restaurants-8k} and the single-domain \textsc{dstc8} datasets, GenSF achieves state-of-the-art results and outperforms prior work. In few-shot settings, we achieve a $30$ $F_1$ score improvement over Span-BERT and Span-ConveRT. On average, GenSF moderately outperforms ConVEx, with $> 2.0$ $F_1$ score improvements in the few-shot settings on \textsc{restaurant-8k}, and both the full data and few-shot settings on two of the \textsc{dstc8} datasets. These experiments empirically validate (1) the importance of aligning the pre-trained model and the downstream task by simultaneously incorporating inductive biases into both the task and the model and (2) that through response generation pre-training, dialog models have implicitly learned to detect certain slots, which can be leveraged by effectively adapting the downstream task.

\subsection{Zero-shot Slot Filling}

\begin{table}[]
\renewcommand*{\arraystretch}{1.1}
\small
    \centering
{%
\begin{tabularx}{0.5\textwidth}{l *{4}{c}}
\toprule

       \textbf{Slot} & \textbf{Metric} & \textbf{Coach+TR} & \textbf{ConVEx} &  \textbf{GenSF}\\ \midrule
  
& P &  1.7 & 2.3 & \textbf{13.7} \\
First Name & R & 4.1 & 20.1  &  \textbf{36.1} \\
& $F_1$ & 2.5 & 4.1  &  \textbf{19.8} \\ \midrule
  
& P & 0 & 1.9  & \textbf{10.6} \\
Last Name & R & 0 & 16.2  & \textbf{19.7}\\
& $F_1$ & 0 & 3.4 & \textbf{13.8}  \\ \midrule
  
& P & 10.2 & 2.2  & \textbf{10.7} \\
Date & R & \textbf{34.8} & 10.1 & 15.3 \\
& $F_1$ & \textbf{15.7} & 3.6 & 12.6 \\ \midrule

& P &  \textbf{47.4} & 5.6 & 27.5 \\
Time & R & 27.9 & 23.6 & \textbf{46.9} \\
& $F_1$ & \textbf{35.1} & 9.1 & 34.7 \\ \midrule
  
& P & 0 & 3.8  &  \textbf{14.5}\\
People & R & 0 & 13.9 & \textbf{18.9} \\
& $F_1$ & 0 & 6.0 &  \textbf{16.4} \\ \midrule \midrule

Average & $F_1$ & 10.7 & 5.2 &  \textbf{19.5} \\ \midrule

\end{tabularx}
}
    \caption{Zero-shot slot filling results on \textsc{restaurants-8k}. All models are evaluated on the test set without any training on the dataset.}
    \label{tab:zero_results}
\end{table}

\begin{table*}[]
\renewcommand*{\arraystretch}{1.1}
\small
    \centering
{%
\begin{tabularx}{0.7\textwidth}{@{\extracolsep{\stretch{1}}} l *{3}{c}}
\toprule

       \textbf{Model} & \textbf{Full-Data} & \textbf{Few-Shot ($\nicefrac{1}{16}$)} &  \textbf{Zero-Shot}\\ \midrule

GenSF & \textbf{96.1} & \textbf{89.7} & \textbf{19.5} \\ \midrule

\multicolumn{4}{c}{Removing Model Adaptation} \\ \midrule

-- Copy-mechanism & 95.6 & 87.8 & \textbf{19.5} \\
-- Constrained Decoding &  95.4 & 89.5 & 0.5     \\
-- Post-processing & \textbf{96.1} & \textbf{89.7}  & 18.1   \\ 
-- All model adaptation & 95.4 & 87.8 & 0.5 \\ \midrule
  
\multicolumn{4}{c}{Removing Task Adaptation} \\ \midrule

-- Natural Language Slot Names & 95.3 & 86.6 & 12.2   \\
-- Natural Language Templates &  94.8 & 88.5 & 0.0 \\
-- All Natural Language & 95.5 & 88.9 &  0.0  \\ \midrule

\multicolumn{4}{c}{Removing All Adaptation} \\ \midrule

-- All Adaptation & 95.8  & 89.2 &  0.0 \\ \bottomrule

\end{tabularx}
}
    \caption{Ablation experiments. We remove (1) adaptations to the model, (2) adaptations to the downstream task and (3) all adaptations proposed in this paper. The experiments are carried out on the full-data, few-shot ($\nicefrac{1}{16}$th of the training set) and zero-shot settings of \textsc{restaurants-8k}.}
    \label{tab:ablation_results}
\end{table*}

For zero-shot slot filling, we must have strong alignment between the pre-trained model and the downstream task. Since the model is not fine-tuned on the task, it is necessary to effectively align the formulation of the downstream task to the capabilities of the model. As such, zero-shot experiments validate our proposed reformulation of slot filling as natural language response generation.

For these experiments, we compare to the published results of ConVEx \citep{henderson2020convex}. Furthermore, we run a Coach+TR model \citep{liu2020coach} on the \textsc{restaurant-8k} dataset. Note that while ConVEx and \textsc{GenSF} have only been trained on open-domain dialog, Coach+TR trains on adjacent task-oriented domains (i.e., SNIPS), meaning that the zero-shot performance is higher on slots that are domain agnostic.

The experiments used the \textsc{restaurants-8k} dataset with the \textsc{GenSF} model. The copy-mechanism is removed from the model, as it adds additional weights to the model and therefore requires training. However, the constrained decoding and the post-processing heuristic of \textsc{GenSF}, allow us to enforce that the slot values will always be a contiguous span from the input utterance. Table \ref{tab:zero_results} demonstrates that \textsc{GenSF} significantly outperforms prior work on zero-shot slot filling with a \textbf{14 $\mathbf{F_1}$ score improvement} over ConVEx and a \textbf{9 $\mathbf{F_1}$ score improvement} over Coach+TR. These results further validate the hypothesis that pre-trained dialog models have implicitly learned to detect slots and that this ability can be leveraged through the proposed task reformulation. 

Most noteworthy is the performance on the \textit{`first name'} and \textit{`last name'} slots. This suggests that, to some degree, DialoGPT \citep{zhang-etal-2020-dialogpt} can disambiguate between a first name and a last name when provided simultaneously (e.g., \textit{`my name is Lakesha Mocher'}). It should be noted that the macro-averaged $F_1$ score used to evaluate the models considers a slot value to be incorrect unless it exactly predicts the ground-truth slot value. In many cases, the \textsc{GenSF} model produces appropriate slot values that differ from the ground-truth, e.g., \textit{`wednesday'} instead of \textit{`next wednesday'}. It is possible that by incorporating additional inductive biases about the specific formulation of the slot values (e.g., slots should have maximal information) into the inference algorithm, the zero-shot performance can be further increased.

\textsc{GenSF} is shown to strongly outperform prior work on zero-shot slot filling. This impressive performance validates the proposed approach of simultaneously adapting both the downstream task and the pre-trained model. Furthermore, zero-shot performance also confirms the hypothesis that pre-trained response generation models have implicitly learned to understand and detect slots, thereby highlighting the potential of leveraging generative pre-trained models for language understanding tasks. Future work should explore mechanisms for reformulating other downstream tasks (e.g., intent prediction, dialog state tracking) in order to leverage generative pre-trained models. Furthermore, it is possible that these zero-shot results could be further improved through two-stage pre-training (e.g., further pre-train with the \textit{`pairwise cloze'} task).

\subsection{Ablation}

\textsc{GenSF} has been shown to outperform prior work in full-data, few-shot and zero-shot settings. To determine the source of the improvements, we perform an ablation study. The ablation experiments remove the adaptations used in \textsc{GenSF} and evaluate on \textsc{restaurants-8k} across full-data, few-shot ($\nicefrac{1}{16}$ of the training set) and zero-shot settings. Removing all the ablation, is equivalent to training a DialoGPT model from scratch on the task, similar to the approach proposed by \citet{madotto2020language}. 

As shown in Table \ref{tab:ablation_results}, the various adaptations are vital to the strong performance of \textsc{GenSF}. Of the model adaptations, only the copy-mechanism is necessary in the full-data setting, since the model effectively learns to copy tokens from the input utterance and therefore does not need constrained decoding and post-processing. However, constrained decoding is necessary for the zero-shot settings, as the zero-shot model does not leverage a copy-mechanism. Task adaptation, especially the use of natural language templates, is shown to be important across all of the experimental settings. This highlights the importance of formulating the downstream task in a manner that can effectively leverage the capabilities of the pre-trained models. 

The results of the ablation study further validate this paper's primary hypothesis. Pre-trained models work better for downstream tasks, when the task and the model are effectively aligned. As shown in the results of the ablation study, removing this adaptation results in a performance decrease.

\section{Conclusion}

This paper simultaneously adapts both the task and the pre-trained model in order to achieve strong alignment between a generative pre-trained dialog model and the downstream slot filling task. The resulting \textsc{GenSF} model achieves state-of-the-art results on two slot filling datasets, with particularly strong gains in few-shot and zero-shot settings. The empirical results underlie the importance of incorporating inductive bias into both the task and the pre-trained model. While this paper demonstrates the value of simultaneous adaptation for the task of slot filling, a similar paradigm could potentially be extended to alternate tasks. Future work should (1) explore improved mechanism for achieving stronger alignment between the task and the model, (2) extend the simultaneous adaptation strategy to other problems and (3) explore the use of pre-trained generative models for language understanding tasks.

\bibliographystyle{acl_natbib}
\bibliography{anthology,acl2021}


\end{document}